\documentclass{article}

     \usepackage[final,nonatbib]{neurips_2024}

\usepackage[utf8]{inputenc} 
\usepackage[T1]{fontenc}    
\usepackage{hyperref}       
\usepackage{url}            
\usepackage{booktabs}       
\usepackage{amsfonts}       
\usepackage{nicefrac}       
\usepackage{microtype}      
\usepackage{xcolor}         
\usepackage{graphicx}
\usepackage{float}
\usepackage{amsmath}
\usepackage{bm}
\usepackage{multirow}

\title{Explaining Chest X-ray Pathology Models using Textual Concepts}

\author{%
  Vijay Sadashivaiah\\
  Department of Computer Science\\
  Rensselaer Polytechnic Institute\\
  Troy, NY 12180 \\
  \texttt{sadasv2@rpi.edu} \\
  \And
  Pingkun Yan\\
  Department of Biomedical Engineering\\
  Rensselaer Polytechnic Institute\\
  Troy, NY 12180 \\
  \texttt{yanp2@rpi.edu} \\
   \AND
  James A. Hendler\\
  Department of Computer Science\\
  Rensselaer Polytechnic Institute\\
  Troy, NY 12180 \\
  \texttt{hendler@cs.rpi.edu} 
}

\begin{document}

\maketitle

\begin{abstract}
Deep learning models have revolutionized medical imaging and diagnostics, yet their opaque nature poses challenges for clinical adoption and trust. Amongst approaches to improve model interpretability, concept-based explanations aim to provide concise and human-understandable explanations of any arbitrary classifier. However, such methods usually require a large amount of manually collected data with concept annotation, which is often scarce in the medical domain. In this paper, we propose Conceptual Counterfactual Explanations for Chest X-ray (CoCoX), which leverages the joint embedding space of an existing vision-language model (VLM)  to explain black-box classifier outcomes without the need for annotated datasets. Specifically, we utilize textual concepts derived from chest radiography reports and a pre-trained chest radiography-based VLM to explain three common cardiothoracic pathologies. We demonstrate that the explanations generated by our method are semantically meaningful and faithful to underlying pathologies.
\end{abstract}

\section{Introduction}

Over the past decade, deep learning (DL) has achieved unprecedented performance in medical imaging-based diagnosis \cite{litjens2017survey}. This success is partly attributed to the increasing availability of data and the rapid development of deep neural network (DNN) technologies. Despite these impressive results, DNNs have been criticized for their black-box nature, which has limited the adoption and regulatory approval of medical imaging-based DNNs \cite{langlotz2019roadmap}. The combination of lack of explainability (challenging to interpret their inner workings) \cite{singh2020explainable}, lack of robustness (being easily fooled using adversarial data) \cite{zhang_overlooked_2022}, bias (tendency to amplify inequalities that exist in data) \cite{obermeyer2019dissecting}, and the high stakes nature of clinical applications \cite{mozaffari2014systematic,vayena2018machine} prevents deployment of black-box models in clinical practice. Explainable artificial intelligence (XAI) methods aim to tackle this by making the decisions and processes of black-box systems understandable and interpretable to humans.

In the medical imaging field, heatmap-style explanations such as saliency maps have been extensively studied to interpret the decisions made by a classifier \cite{simonyan2013deep,selvaraju2017grad}. However, simply highlighting the relevant pixels that influence classifiers' decisions does not answer {\em why} that region is important\cite{ghassemi2021false}. Additionally, previous work has shown that these methods are susceptible to adversarial attacks \cite{zhang2023revisiting,gu2019saliency}, prone to confirmation bias \cite{adebayo2018sanity,bornstein2016artificial} and require domain expert's intervention to {\em interpret} the explanation leading to subjective bias \cite{kim2018interpretability}. 

On the other hand, concept-based explanation methods provide human-understandable and high-level semantically meaningful explanations \cite{kim2018interpretability,ghorbani2019towards}. In natural images, a {\em concept} represents any semantically meaningful attribute and is typically represented by words such as ``stripes'', ``smile'', etc. To measure the importance of such concepts on the output of a black-box classifier, concept activation vector (CAV) \cite{kim2018interpretability} was proposed. A CAV measures the direction of each concept in the embedding space of a pre-trained classifier. Using CAVs, one can then calculate which concepts influenced the output of a particular class. However, in order to train CAVs, one needs human-annotated concept labels for images, which can be expensive and often infeasible. Another line of work aims to explain pre-trained classifiers using counterfactual explanations \cite{goyal2019counterfactual,stepin2021survey}. In simple terms, counterfactual explanation means identifying the feature perturbations required in input images to lead the pre-trained model to a different output prediction. Few methods have explored this line of work in medical imaging \cite{atad2022chexplaining,singla_explaining_2023}. A key limitation of this approach is the need for subject matter experts to evaluate and assign meaning to generated counterfactual images\cite{atad2022chexplaining}. Additionally, generating counterfactual images requires training GAN-based methods that can capture latent representations from the input data. Training these models can be expensive \cite{karras2020analyzing} and often infeasible due to limited data availability in medical domains \cite{bowles2018gan}. While \cite{kayser2022explaining} introduced a natural language explanation (NLE) dataset derived from MIMIC-CXR, which involved extracting explanations for different pathologies from radiology reports and suggesting retraining black-box classifiers to align with these NLEs, we propose a different approach. Our method focuses on explaining black-box classifiers without the need for retraining by utilizing existing vision-language models (VLMs).

To tackle this challenge, we propose to develop a conceptual counterfactual explanation (CoCoX) that combines the benefits of concept-based and counterfactual explanations.
Specifically, we gathered chest radiology-related natural language concepts from the PadChest dataset \cite{bustos2020padchest}, which is annotated by radiologists and mapped against the Unified Medical Language System (UMLS) database for validity and augmented additional concepts by querying ChatGPT \cite{chatgpt2024} for pathology-specific visual attributes. We then leverage the joint embedding space of a pre-trained vision-language model (VLM) to derive concept directions and create a {\em concept bank}. Next, we train simple projection models that allow us to move from the black-box classifier's latent space to VLM's latent space. Finally, we generate {\em conceptual counterfactuals} by learning perturbations in the direction of each concept in our concept bank. In essence, we perturb the projected image embedding such that the black-box model's prediction changes to the target class, similar to a previous method CounTEX \cite{kim2023grounding}. 

\begin{figure}[t]
\centering
\includegraphics[width=\textwidth]{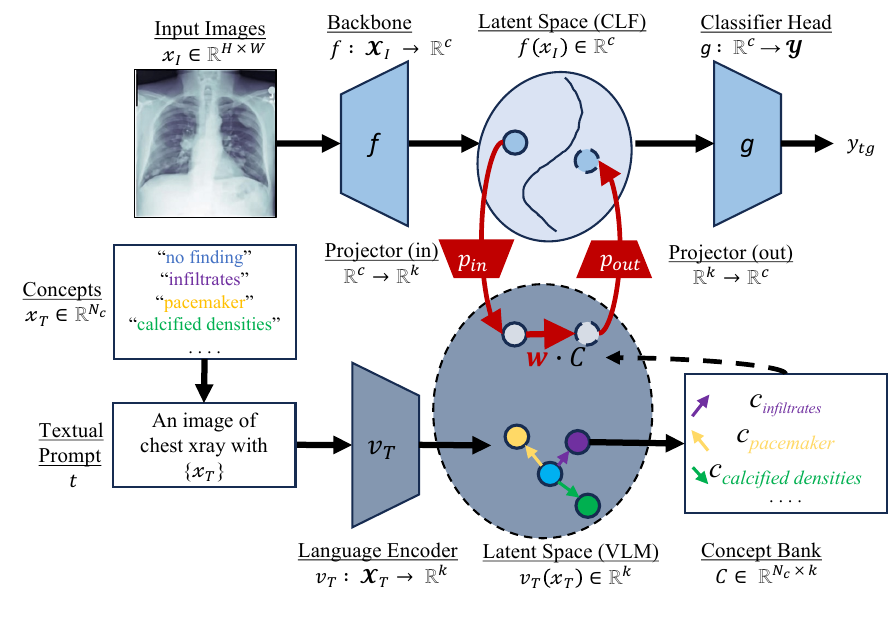}
\caption{The overall architecture of CoCoX. We create the Concept Bank ($C$) by encoding natural language concepts with a chest radiography pre-trained language encoder of a VLM. We then learn Projector(in) and Projector(out) functions $p_{in}$ and $p_{out}$ using MLPs to transfer feature representation between CLF latent space and VLM latent space. Additionally, we learn perturbation weights for each concept in our concept bank, which serves as an importance score in changing the target classifier output. Learnable parameters are highlighted in red arrows.} \label{fig1}
\end{figure}

Given an input image, an actual prediction by a black-box model, the target prediction, and a concept bank, our approach answers the question of {\em how much} should each concept be perturbed to change the output from current prediction to the target prediction? To achieve this, we learn perturbation weights for each concept in our concept bank, which serves as an importance score for each concept in changing the target classifier output. We present CoCoX's overall model architecture in Figure \ref{fig1}. The main contributions of this paper are summarized as follows: 1) we propose a conceptual counterfactual explanation method for chest radiography image classifiers that integrate the benefits of concept-based explanations and counterfactual explanations, and 2) we conduct in-depth experimental analysis to demonstrate that our proposed approach provides clinically relevant conceptual explanation for several chest x-ray pathologies.

\section{Method}
Our proposed method can be divided into three stages: 1) Construction of concept bank, 2) Learning Projector functions, and 3) Learning conceptual perturbations. We discuss each of these steps in more detail below.

\subsection{Constructing concept bank}
We construct our concept bank ($C$) using natural language concepts derived from PadChest dataset \cite{bustos2020padchest}. First, for each textual concept $x_T$, we generate a pair of text prompts composed of a {\em neutral} sequence ($t_{n}$) and a {\em stimuli} sequence ($t_s$).
The neutral sequence is fixed for all concepts with a generic phrase ``\texttt{An image of chest xray with No Finding}'', following the zero-shot prompt strategy from the original CLIP paper \cite{radford2021learning}. Stimuli sequence is generated using the phrase ``\texttt{An image of chest xray with \{$x_T$\}}'', for each concept $x_T$. We then use the pair of prompts $[t_n, t_s]$ to derive the concept direction. First, the pair is tokenized and encoded using the pre-trained language encoder ($v_T$), yielding the embedded prompt pair $[v_T(t_n), v_T(t_s)]$. Then, the direction of textual concept $c \in C$ can be obtained by taking the difference between these two embedded prompt pairs and normalizing it to a unit vector, i.e.,
\begin{equation}
c = \frac{v_T(t_s) - v_T(t_n)}{||v_T(t_s) - v_T(t_n)||_2}.
\label{eq:concept}
\end{equation}
We compute a textual concept $c \in \mathbb{R}^{k}$ for each stimulus to obtain our final concept bank $C = \{c_i | i=1,\ldots, N_c \}$, where $k$ is the embedding dimension of VLP and $N_c$ is the number of concepts. 

\subsection{Learning projection functions}
\begin{figure}[t]
\centering
\includegraphics[width=0.7 \textwidth]{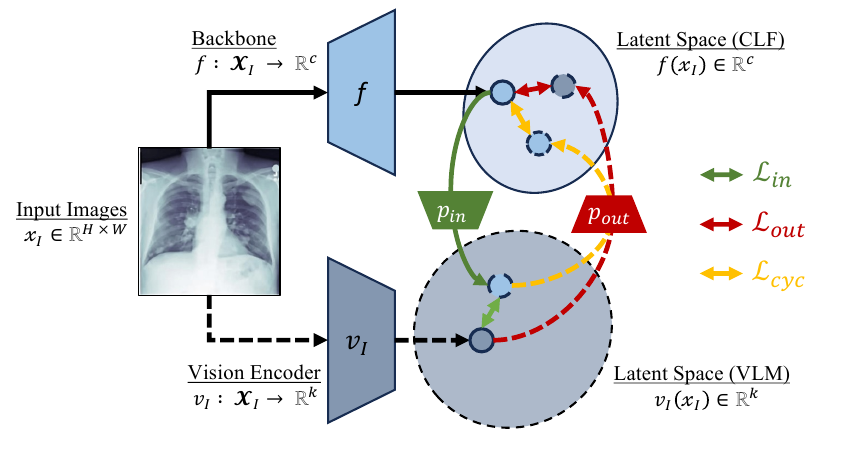}
\caption{Training paradigm for projection functions. Each $\leftrightarrow$ indicates a distance minimization problem. Total loss is a sum of all loss functions, $\mathcal{L}_{total} = \mathcal{L}_{{in}} + \mathcal{L}_{{out}} + \mathcal{L}_{{cyc}}$.}
\label{loss}
\end{figure}

Since our goal is to operate on textual concepts encoded in VLM latent space, we need to learn the projections $p_{in}$ and $p_{out}$, that can move image embedding from classifier (CLF) latent space to VLM latent space and vice versa. First, to train the $p_{in}$ function, we project the input embedding $f(x_I)$ into VLM latent space resulting in the projected embedding $p_{in}(f(x_I))$. Next, we embed the same image using the vision encoder $v_I$ of VLM to obtain $v_I(x_I)$. Ideally, we want these two embeddings to be close to each other, and hence, the $p_{in}$ is optimized to minimize this distance. Similarly, for $p_{out}$, we first project the VLM vision encoder's embedding ($v_I(x_I)$) to classifier latent space using $p_{out}$, resulting in the projected embedding $p_{out}(v_I(x_I))$. We also encode the same image using the pre-trained classifier, which results in embedding $f(x_I)$. Again, $p_{out}$ is trained to minimize the distance between these two embeddings. 
Finally, we want to account for the round-trip error that minimizes the distance between embedding $p_{out}(p_{in}(f(x_I)))$ and $f(x_I)$. Taking all these three into account, we get the following loss function,
\begin{equation}
    \begin{split}
        \mathcal{L}_{total} & = \mathcal{L}_{{in}} + \mathcal{L}_{{out}} + \mathcal{L}_{{cyc}} \\
                            & = ||p_{in}(f(x_I)) - v_I(x_I)||^2 + ||p_{out}(v_I(x_I)) - f(x_I)||^2 \\
                            & \quad + ||p_{out}(p_{in}(f(x_I))) - f(x_I)||^2
    \end{split}
\label{eq:projector}
\end{equation}
We pre-train $\mathcal{L}_{{in}}$ and $\mathcal{L}_{{out}}$ independently and fine-tune $\mathcal{L}_{total}$ for a few epochs. See Figure \ref{loss} for a depiction of this training paradigm.

\subsection{Learning conceptual perturbations}
Once the concept bank is established and the projection functions are trained, we need to learn concept-wise perturbation weight parameter $\bm{w}$, such that the gap between perturbed prediction $y_p$ and target prediction $y_t$ is minimized. Since we apply this perturbation in VLM latent space, we first project our input image $x_I$ to VLM latent space using $p_{in}$, apply learnable perturbation, project back to classifier latent space using $p_{out}$, and predict the output of perturbed embedding. 
We define this output below
\begin{equation}
    y_p = g(p_{out}(p_{in}(f(x_I)) + \bm{w} \cdot C)),
\end{equation}
where $g$ is the classifier head of our pre-trained classifier. 
In order to optimize $\bm{w}$, we introduce the following constraints. First, the perturbed embedding should predict the target class $y_t$. We propose using cross-entropy loss between the perturbed and target classes, $\mathcal{L}_{CE} = CE((y_p, y_t))$. 
Next, we introduce sparseness through L1 regularization $||\bm{w}||_1$, and ensure that the perturbed embedding does not deviate from image embedding using L2 regularization $||\bm{w}||_2$. Our final loss $\mathcal{L}_{final}$ is a weighted sum of these three loss functions
\begin{equation}
    \mathcal{L}_{final} = \mathcal{L}_{CE} + \alpha ||\bm{w}||_1 + \beta ||\bm{w}||_2,
    \label{eq:perturbation}
\end{equation}
where $\alpha$ and $\beta$ are weighting parameters.

\section{Experiments}
In this section, we describe experimental details, including datasets, implementation, and metrics. We also present the quantitative and qualitative results, comparing the explanations generated by CoCoX against clinical literature and an existing baseline.

\subsection{Datasets}
{\bfseries Imaging datasets:} We used two large-scale chest x-ray image datasets in our experiments. First, we used the CheXpert dataset \cite{irvin2019chexpert} to train our pathology backbone classifiers. Following the previous work \cite{atad2022chexplaining}, we trained for three different pathologies, Cardiomegaly, Atelectasis, and Pleural Effusion. Next, to train our projector models, we used the MIMIC-CXR dataset \cite{johnson2019mimic}.

{\noindent \bfseries Concept bank:} To create our concept bank $C$, we started with the radiological findings annotated in the PadChest dataset \cite{bustos2020padchest}. This yielded 174 natural language concepts. Next, we queried ChatGPT \cite{chatgpt2024} with prompt ``\texttt{What visual attributes are seen in $pathology$ chest xray?}'' for each of the pathologies and augmented our concept library yielding a total of 192 concepts\footnote{See Supplementary Section 2 for an example query and response}. It is important to note the ease of augmenting additional concepts to the library. CoCoX allows the addition/removal of textual concepts, unlike other concept-based explanation methods that need annotated images. We then proceed with Eq.~\ref{eq:concept} to create the concept bank.

\subsection{Implementation details}
{\noindent \bfseries Backbone classifiers and VLM: } We implemented three different architectures for our backbone image classifiers. Specifically, we chose DenseNet121, ResNet34, and VLM+Linear for our experiments. For VLM+Linear based architecture, we just added a linear layer on top of our VLM vision encoder (ViT-B/32). Unlike the other two architectures, the VLM-based one does not require separate projector models since are already operating in VLM latent space. Each model was trained for pathologies using the CheXpert dataset for a maximum of 50 epochs with early stopping. For our VLM, we used CheXzero model \cite{tiu2022expert}.

{\noindent \bfseries Projectors: }
We used a simple 2-layer multi-layer perceptron (MLP) to implement $p_{in}$ and $p_{out}$ functions. Both these models comprised of (512, 512) hidden units and were trained with the MIMIC-CXR dataset for a maximum of 50 epochs with early stopping to minimize the loss function in Equation \ref{eq:projector}. A batch size of 64 was used for all the experiments.

{\noindent \bfseries Conceptual perturbation: } 
The weight parameter $\bm{w}$ is optimized to minimize the loss $\mathcal{L}_{final}$ (see Eq.~\ref{eq:perturbation}). Parameters $\alpha$ and $\beta$ are set to 0.1 for all experiments following \cite{kim2023grounding}. We used stochastic gradient descent (SGD) with momentum and a learning rate of $10^{-2}$ for a maximum of 100 steps. We stopped training early if the prediction of input image changes to the desired target class. This weight parameter serves as an importance score for each concept in changing the output of the black box classifier to the target output.

\subsection{Results and Evaluation}

For each of the three pathologies, Cardiomegaly, Pleural Effusion, and Atelectasis, we optimize the perturbations to change a ``No Finding'' image to corresponding pathology finding. To verify that the proposed approach works as intended, we conducted an experiment where we {\em included} the pathology name as one of the concepts in our concept bank. After perturbing 100 random example images from each pathology to the target class using our VLM + Linear model, we found that the pathology name was \textbf{top@1} concept for 99\% of Cardiomegaly, 100\% of Pleural Effusion and 98\% of Atelectasis cases. The corresponding pathology name is removed from the concept bank for the next set of experiments.

We present random images for each pathology and corresponding top 5 conceptual counterfactuals generated by our model for VLM + Linear and DenseNet121 model in Figure \ref{fig:examples}. For each pathology, the emerging top concepts represent the main features of each disease. For instance, the top concepts in converting a ``No Finding'' image to ``Cardiomegaly'' are \texttt{increased cardiac diameter, increased cardio-thorasic ratio}, correspond to prominent features in Cardiomegaly\cite{amin2021cardiomegaly}. For Pleural Effusion, our method picks up relevant features such as \texttt{increased opacity in lower lungs, fluid accumulation} in the top 5 concepts\cite{krishna2023effusion}. In essence, these concepts had the most impact when {\em added} to input images to change from ``No Finding'' to a pathology finding. Similarly, it is also relevant to analyze conceptual counterfactuals for changing pathology input to ``No Finding'' output. We present this result in Figure S1 for all pathologies. Additionally, we observe alignment between top concepts chosen by DenseNet121 and VLM+Linear model, demonstrating the efficacy of our approach on various model architectures. 

\begin{figure}[t]
\centering
\includegraphics[width=1.0\textwidth]{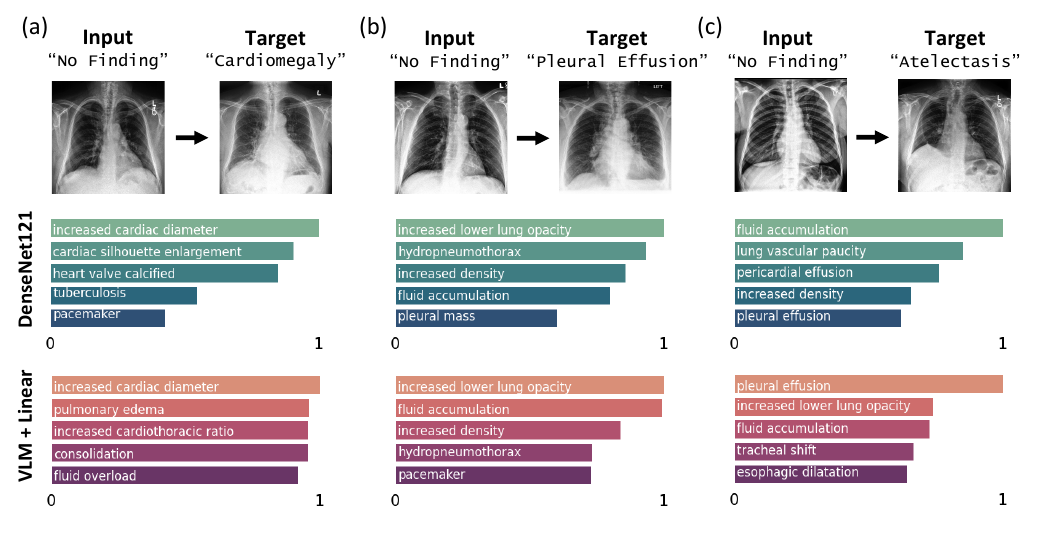}
\caption{\textbf{Conceptual counterfactual generated by CoCoX.} The input image into the model for each pathology is shown under the heading Input (``No Finding''). The target image presented under Target (``Pathology'') is a representative image labeled positive for the corresponding pathology. The concept importance scores for the top 5 concepts are visualized with blue-green for the DenseNet-121 model and orange-purple for VLM + Linear. These concepts had the most impact when added to input images to change from ``No Finding'' to a pathology finding.} 
\label{fig:examples}
\end{figure}

In order to validate that the generated conceptual counterfactuals are clinically relevant, we compare our method against CheXplaining in Style \cite{atad2022chexplaining}. The authors used conditional StyleGAN \cite{karras2020analyzing} based architecture to generate counterfactual images for three chest x-ray pathologies. In their evaluation, the radiologists first listed the primary features and possible secondary findings they rely on while diagnosing each disease. Then, the authors perturbed ten random ``No Finding'' samples to generate counterfactual images, which radiologists evaluated and annotated for the presence or absence of primary and secondary features. However, the authors did not present the fraction (out of ten) of counterfactuals that contained these features. In our experiment, we first verified if the features relied on by radiologists were already in our concept bank. If a feature was absent, we appended that to our concept bank. Next, we calculate the overlap between features identified by radiologists and concepts identified by our approach. We use the recall@k measure typically used in recommender systems evaluation. Recall@k measures the fraction of ground-truth concepts (i.e., radiologists' list) present in top@k predictions made by our model. We present mean recall@k results for all three models in Table \ref{tab:topk} using 1000 images, where we separate the scores by primary and secondary findings\footnote{see Supplementary Table S1 for the list of primary and secondary findings}. We also list the number of primary and secondary features annotated by radiologists within the parentheses for each pathology. Table \ref{tab:topk} shows that our model can recall the primary findings for all three pathologies with high recall scores. 
Recall@k measure penalizes more when the ground truth list is extensive, which is the case for secondary features. Hence, we observe a lower recall@5 values for secondary features. Additionally, concepts such as ``Older patients'' in the secondary list for Cardiomegaly and Atelectasis are non-visual attributes and, hence, challenging to generate conceptual counterfactuals. We bold the highest score in each row.

\setlength{\tabcolsep}{4pt}
\begin{table}[t]
\centering
\caption{Comparison of recall (R) at k concepts to radiologists' evaluation.}
\begin{tabular}{llcccccc}
\toprule
\multirow{2}{*}{\textbf{\begin{tabular}[c]{@{}l@{}}Pathology\end{tabular}}} & \multirow{2}{*}{\textbf{Finding}} & \multicolumn{2}{c}{\textbf{DenseNet121}} & \multicolumn{2}{c}{\textbf{VLM + Linear}} & \multicolumn{2}{c}{\textbf{ResNet34}}  \\ \cmidrule(lr){3-4} \cmidrule(lr){5-6} \cmidrule(lr){7-8} 
                                    &                     & R@5   & R@10  & R@5   & R@10  & R@5  & R@10     \\ \midrule
\multirow{2}{*}{Cardiomegaly}       & {\em Primary(1)}    & 0.97  & \textbf{1.00}  & \textbf{1.00}  & \textbf{1.00}  & 0.98 & \textbf{1.00}     \\         
                                    & {\em Secondary(4)}  & 0.35  & 0.48  & 0.50  & 0.50  & 0.32 & \textbf{0.52}     \\ \cmidrule{2-8}
\multirow{2}{*}{Pleural Effusion}   & {\em Primary(2)}    & 0.46  & 0.78  & 0.50  & \textbf{1.00}  & 0.43 & 0.81     \\
                                    & {\em Secondary(3)}  & 0.17  & 0.29  & \textbf{0.33}  & 0.33  & 0.19 & 0.25     \\ \cmidrule{2-8}
\multirow{2}{*}{Atelectasis}        & {\em Primary(2)}    & 0.55  & 0.69  & 0.50  & \textbf{0.85}  & 0.48 & 0.73     \\
                                    & {\em Secondary(3)}  & 0.26  & 0.61  & 0.33  & \textbf{0.66}  & 0.21 & 0.59     \\ \bottomrule
\end{tabular}
\label{tab:topk}
\end{table}

Finally, we demonstrate that our model generates conceptual counterfactuals faster than \cite{atad2022chexplaining}. Our model using VLM+Linear generated explanations for 99\% of input images (n=600 images) in under 30 seconds, whereas CheXplaining in Style takes 5 minutes to create explanations with an average of 94\% coverage.

\section{Conclusions}
In order to improve the explainability of black-box medical-image classifiers, we explored the chest x-ray domain and investigated conceptual counterfactual explanations that combine concept-based and counterfactual explanations. Our approach highlights important concepts that contribute to changing an image from ``No Finding'' to a pathology finding and vice-versa. We created a concept bank using radiological findings from the PadChest \cite{bustos2020padchest} dataset and augmented additional concepts by querying ChatGPT \cite{chatgpt2024}. Next, to generate conceptual counterfactual explanations, we manipulate the latent embedding of input image with perturbations, such that the classifier's output changes to target class. We found that the top concepts generated by our method are relevant to the underlying pathology and align with radiologists' evaluation. While we have qualitatively demonstrated clinical relevance by aligning with radiologists' assessments, we plan to explore standardized metrics like expert consensus scores, and counterfactual faithfulness in future iterations.

With the proliferation of vision-language models \cite{tiu2022expert,huang2023visual}, our method can easily be extended to studying black-box classifiers in other medical imaging domains. This extension would involve the careful curation of domain-specific concepts tailored to the target medical imaging modality. For instance, in MRI or CT scans, concepts related to soft tissue abnormalities or bone structure could be incorporated into the concept bank. Additionally, training the CoCoX modules on these curated concepts allows us to generate useful conceptual counterfactual explanations for various medical conditions beyond chest X-rays. We leave this for future work, along with exploring ways to improve concept bank creation.

%
%
%
\bibliographystyle{splncs04}
\bibliography{mybibliography}

%

\section*{Supplementary Materials}

\renewcommand{\thetable}{S1}
\setlength{\tabcolsep}{4pt}
\begin{table}[h!]
    \centering
    \caption{List of primary and secondary features annotated by radiologists for each pathology in CheXplaining in Style \cite{atad2022chexplaining}}
    \begin{tabular}{lll}
\hline
\multirow{2}{*}{\textbf{Pathology}} & \multirow{2}{*}{Type} & \multirow{2}{*}{\textbf{Features}}                                                                                  \\
                                    &                       &                                                                                                                     \\ \hline
\multirow{5}{*}{Cardiomegaly}                        & \textit{Primary}      & Increased cardiothoracic ratio                                                                                      \\ \cline{2-3} 
                                    & \textit{Secondary}    & \begin{tabular}[c]{@{}l@{}}Reduced lung tissue opacity\\ Pleural Effusion\\ Pacemaker\\ Older patients\end{tabular} \\ \hline
\multirow{4}{*}{Pleural Effusion}                    & \textit{Primary}      & \begin{tabular}[c]{@{}l@{}}Obstruction of the pleural recessus\\ Opaque lower lungs\end{tabular}                    \\ \cline{2-3} 
                                    & \textit{Secondary}    & \begin{tabular}[c]{@{}l@{}}Increased cardiac diameter \\ Fluid overload\\ Pneumonia\end{tabular}                    \\ \hline
\multirow{4}{*}{Atelectasis}                         & \textit{Primary}      & \begin{tabular}[c]{@{}l@{}}Mediastinal shift\\ Wide barrel-like thorax\end{tabular}                                 \\ \cline{2-3} 
                                    & \textit{Secondary}    & \begin{tabular}[c]{@{}l@{}}Pleural Effusion\\ Infiltration\\ Older patients\end{tabular}                            \\ \hline
\end{tabular}
    \label{tab:features}
\end{table}

\section*{Example concepts generated by prompting ChatGPT\cite{chatgpt2024}}
\textbf{Prompt:} \texttt{What visual attributes are seen in cardiomegaly chest xray?}\\

\noindent \textbf{Response:}
\texttt{\\
Cardiac silhouette enlargement\\
Increased cardiothoracic ratio\\
Prominent pulmonary vessels\\
Pleural effusion\\
Signs of heart failure such as pulmonary edema}
\newpage
\renewcommand{\thefigure}{S1}
\begin{figure}[htp!]
\centering
\includegraphics[width=1.0\textwidth]{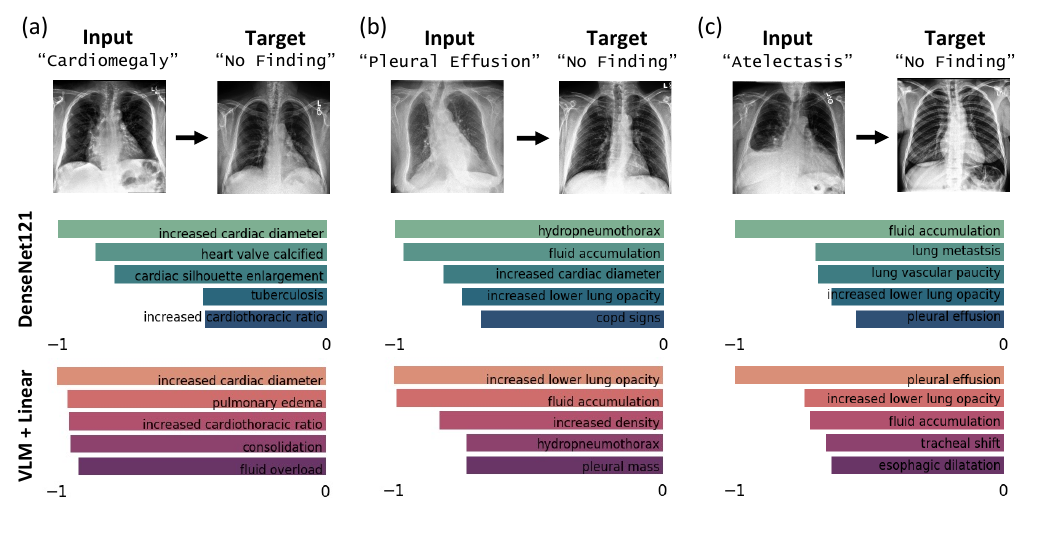}
\caption{\textbf{Reverse conceptual counterfactual generated by CoCoX.} The input image into the model is shown under the heading Input (``Pathology''). The target image presented under Target (``No Finding'') is a representative image labeled negative for any pathologies. The concept importance scores for the top 5 concepts are visualized with blue-green for the DenseNet-121 model and orange-purple for VLM + Linear. These concepts had the most impact when subtracted from the input image embedding to change from pathology finding to no finding.} 
\label{fig:examples}
\end{figure}

\end{document}